\documentclass[10pt,twocolumn,letterpaper]{article}
\pdfoutput=1
\usepackage{cvpr}
\usepackage{times}
\usepackage{epsfig}
\usepackage{graphicx}
\usepackage{amsmath}
\usepackage{amssymb}
\usepackage{enumerate}

\usepackage[pagebackref=true,breaklinks=true,letterpaper=true,colorlinks,bookmarks=false]{hyperref}

\cvprfinalcopy % *** Uncomment this line for the final submission
\DeclareMathOperator*{\argmax}{arg\,max} % Jan Hlavacek
\def\cvprPaperID{1742} % *** Enter the CVPR Paper ID here

\begin{document}
%%%%%%%%% TITLE
\title{Improving Interpretability of Deep Neural Networks with Semantic Information
\thanks{The work was supported by the National Basic Research Program (973 Program) of China (No. 2013CB329403), 
National NSF of China Projects (Nos. 61571261, 61620106010, 61621136008), China Postdoctoral Science Foundation (No. 2015M580099), 
Tiangong Institute for Intelligent Computing, and the Collaborative Projects with Tencent.}}
\author{Yinpeng Dong \hspace{1.2cm} Hang Su \hspace{1.2cm} Jun Zhu \hspace{1.2cm} Bo Zhang \\
Tsinghua National Lab for Information Science and Technology\\
 State Key Lab of Intelligent Technology and Systems\\
Center for Bio-Inspired Computing Research\\
Department of Computer Science and Technology, Tsinghua University\\
\small{dongyinpeng@gmail.com \hspace{0.3cm} \{suhangss, dcszj, dcszb\}@mail.tsinghua.edu.cn}
}

\maketitle
%\thispagestyle{empty}

%%%%%%%%% ABSTRACT
\begin{abstract}
Interpretability of deep neural networks (DNNs) is essential since it enables users to understand the overall strengths and weaknesses of the models, conveys an understanding of how the models will behave in the future, and how to diagnose and correct potential problems. However, it is challenging to reason about what a DNN actually does due to its opaque or black-box nature. To address this issue, we propose a novel technique to improve the interpretability of DNNs by leveraging the rich semantic information embedded in human descriptions. By concentrating on the video captioning task, we first extract a set of semantically meaningful topics from the human descriptions that cover a wide range of visual concepts, and integrate them into the model with an interpretive loss. We then propose a prediction difference maximization algorithm to interpret the learned features of each neuron.
Experimental results demonstrate its effectiveness in video captioning using the interpretable features, which can also be transferred to video action recognition. By clearly understanding the learned features, users can easily revise false predictions via a human-in-the-loop procedure.
\end{abstract}

%%%%%%%%% BODY TEXT
\section{Introduction}

\begin{figure}[t]
\begin{center}
  \includegraphics[width=1.0\linewidth]{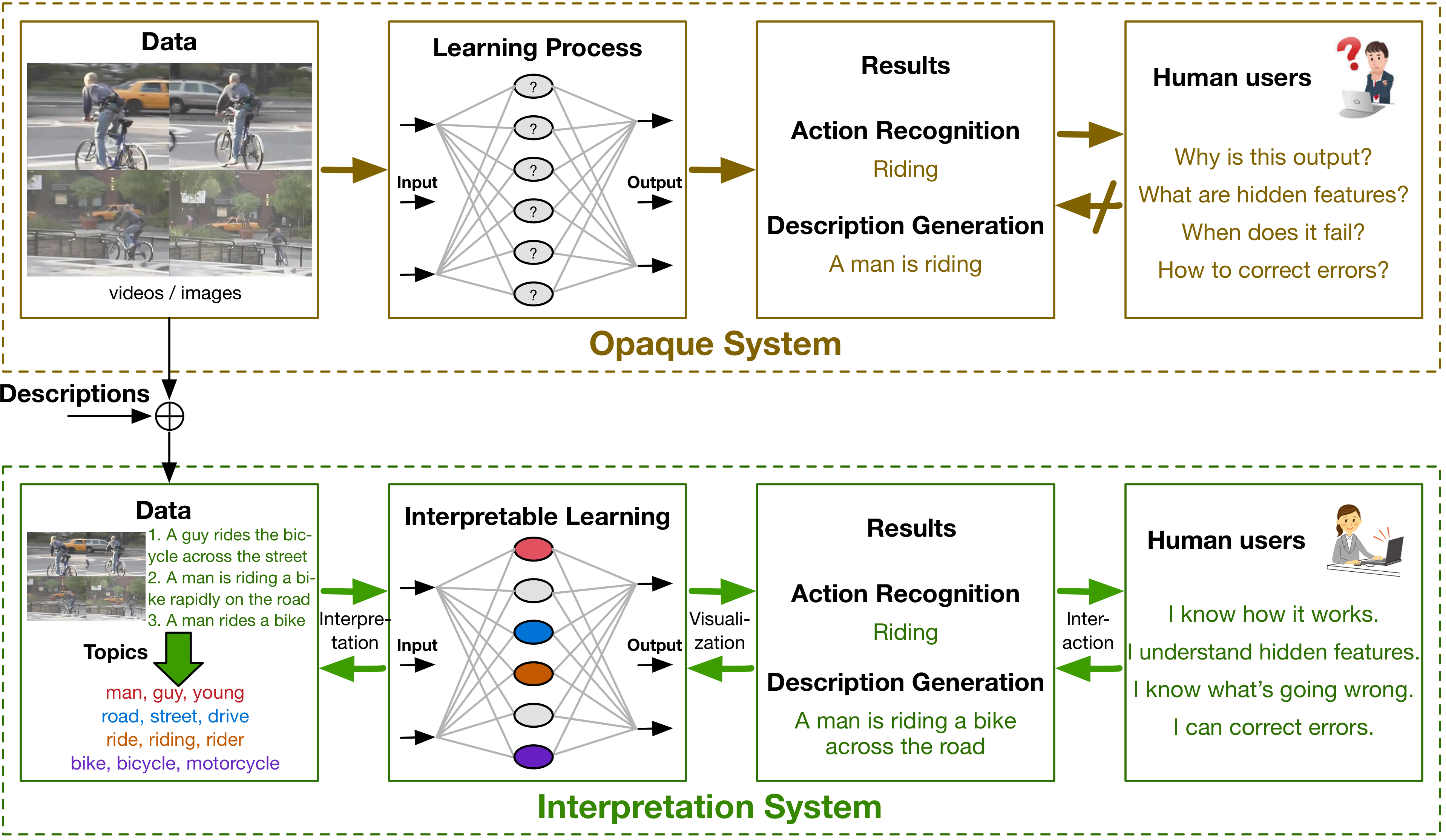}
\end{center}\vspace{-2ex}
   \caption{An overview of our interpretation system (bottom) compared with an opaque system (top). An opaque system often learns abstract and incomprehensible features. Human users have to accept the decisions from the system passively, but are unable to understand the rationale of the decisions and interact with it. To address this issue, we incorporate topics embedded in human descriptions as semantic information, to improve interpretability of DNNs during the learning process. The learned features of each neuron can be associated with a topic (\eg, topic ``road'' with top related words like road, street, and drive can interpret the learned features of the blue neuron). With the aids of these interpretable features, human users can easily visualize and interact with the system, which allows a human-in-the-loop learning procedure.}
\vspace{-2ex}
\label{fig:overview}
\end{figure}

Deep Neural Networks (DNNs) have demonstrated state-of-the-art and sometimes human-competitive performance in numerous vision-related tasks~\cite{DeepReview_Lecun_2015}, including image classification~\cite{krizhevsky2012imagenet,szegedy2015going}, object detection~\cite{girshick2014rich,ren2015faster} and image/video captioning~\cite{xu2015show,yao2015describing}. 
With such success, DNNs have been integrated into various intelligent systems as a key component, \eg., autonomous car~\cite{huval2015empirical,bojarski2016end}, medical image analysis~\cite{Greenspan_2016_TMI}, financial investment~\cite{Fin_Syed_2016}, \etc. 
The high-performance of DNNs highly lies on the fact that they often stack tens of or even hundreds of nonlinear layers, and encode knowledge as numerical weights of various node-to-node connections. 

Although DNNs offer tremendous benefits to various applications, they are often treated as ``black box'' models because of their highly nonlinear functions and unclear working mechanism~\cite{bengio2013representation}. Without a clear understanding of what a given neuron in the complex models has learned and how it interacts with others, the development of better models typically relies on trial-and-error~\cite{zeiler2014visualizing}.
Furthermore, the effectiveness of DNNs is partly limited by its inabilities to explain the reasons behind the decisions or actions to human users. It is far from enough to provide eventual outcomes to the users especially for highly regulated environments, since they may also need to understand the rationale of the decisions. For example, a driver of an autonomous car is eager to recognize why obstacles are reported so that he/she can decide whether to trust it; and radiologists also require a clearly interpretable outcome from the system such that they can integrate the decision with their standard guideline when they make diagnosis. As an extreme case in~\cite{nguyen2015deep}, a DNN can be easily fooled, \ie, it is possible to produce images that DNNs believe to be recognizable objects with nearly certain confidence but are completely unrecognizable to humans. In summary, the counter-intuitive properties and the black-box nature of DNNs make it almost impossible for one to reason about what they do, foreseen what they will do, and fix the errors when potential problems are detected. Therefore, it is imperative to develop systems with good interpretability, which is an essential property for users to clearly understand, appropriately trust, and effectively interact with the systems.

Recently, many research efforts have been devoted to interpreting hidden features of DNNs~\cite{erhan2009visualizing,nguyen2016multifaceted,zhou2014object,zeiler2014visualizing}, and have made several steps towards interpretability, \eg, the de-convolutional networks~\cite{zeiler2014visualizing} to visualize the layers of convolutional networks, and the activation maximization~\cite{erhan2009visualizing} to associate semantic concepts with neurons of a CNN. A few attempts have also been made to explore the effectiveness of various gates and connections of recurrent neural networks (RNNs)~\cite{chung2014empirical,karpathy2015visualizing}. Interpretability also bring us some benefits like weakly supervised detection~\cite{zhou2016learning}. However, these works often focus on analyzing relatively simple architectures such as AlexNet~\cite{krizhevsky2012imagenet} for image classification. There still lack interpretation techniques for more complex architectures that integrates both CNN and RNN, in which the learned features are difficult to interpret and visualize.
More importantly, these methods perform interpretation and visualization after the training process. It means that they can only explain a given model, but are unable to learn an interpretable model. Such a decoupling between learning and interpretation makes it extremely hard (if possible at all) to get humans to interact with the models (\eg, correct errors).

In this paper, we address the above limitations by presenting a method that incorporates the interpretability of hidden features as an essential part during the learning process. A key component of our method is to measure the interpretability and properly regularize the learning. Instead of pursuing a generic solution, we concentrate our attention on the video captioning task~\cite{venugopalan2014translating}, for which DNNs have proven effective on learning highly predictive features while the interpretability remains an issue as other DNNs do. 
In this task, we leverage the provided text descriptions, which include rich information, to guide the learning.
We first extract a set of semantically meaningful topics from the corpus, which cover a wide range of visual concepts including objects, actions, relationships and even the mood or status of objects, therefore suitable to represent semantic information.
Then we parse the descriptions of each video to get a latent topic representation, \ie, a vector in the semantic space. We integrate the topic representation into the training process by introducing an \textit{interpretive loss}, which helps to improve the interpretability of the learned features. 

To further interpret the learned features, we present a \textit{prediction difference maximization} algorithm. We also present a human-in-the-loop learning procedure, through which users can easily revise false predictions and  the model based on the good interpretation of the learned features. Our results on real-world datasets demonstrate the effectiveness. 

\section{Methodology}
\begin{figure*}
\begin{center}
\includegraphics[width=0.98\linewidth]{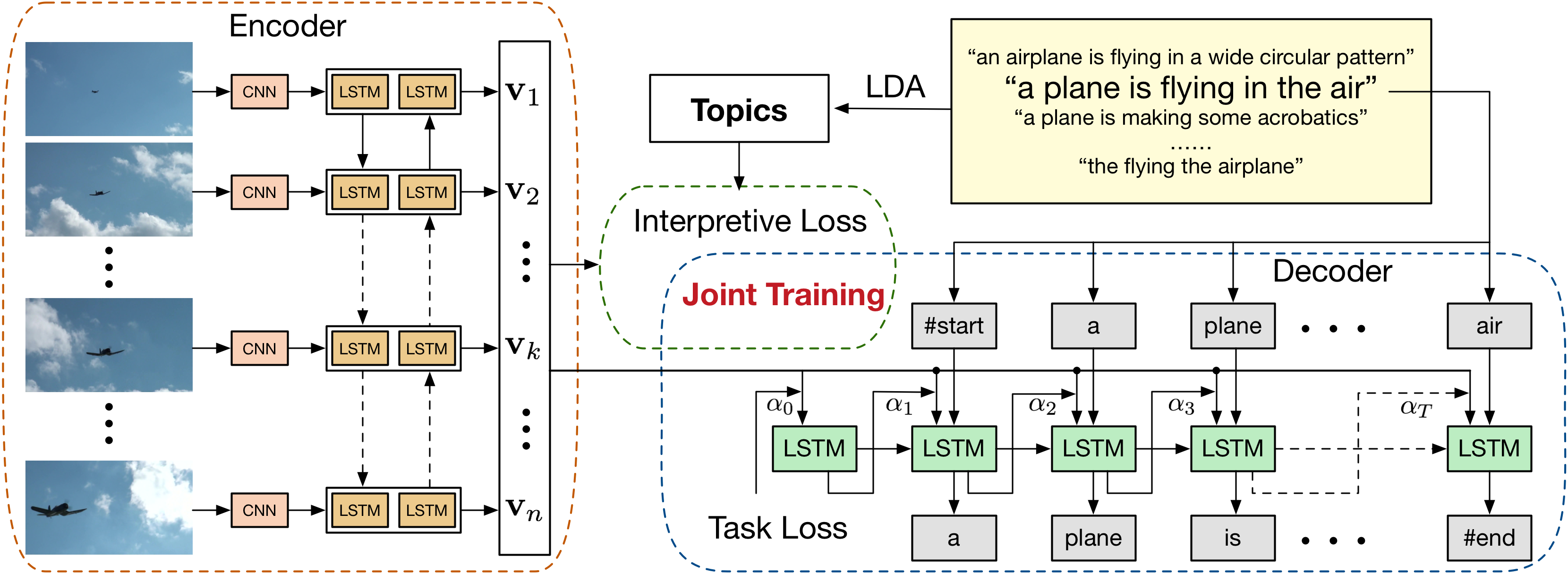}
\end{center}\vspace{-2ex}
   \caption{The attentive encoder-decoder framework for the video captioning task, which can automatically learn interpretable features. We stack a CNN model and a bi-directional LSTM model as encoder to extract video features $\{\mathbf{v}_1,...\mathbf{v}_n\}$, and then feed them to an LSTM decoder to generate descriptions. The attention mechanism is used to let the decoder focus on a weighted sum of temporal features with weight $\alpha_t$. We extract latent topics from human labeled descriptions as semantic information and introduce an interpretive loss to guide the learning towards interpretable features, which is optimized jointly with the negative log-likelihood of training descriptions.
}
   \vspace{-2ex}
\label{fig:framework}
\end{figure*}

In this section, we present the key components of our interpretation system. We first overview the system on the video captioning task. We then present an attentive encoder-decoder network, which incorporates an interpretive loss to learn interpretable features. Afterwards, we  present a prediction difference maximization algorithm to interpret the learned features of each neuron. We will introduce a human-in-the-loop learning procedure by leveraging the interpretability in Section.~\ref{sec:human-in-the-loop}. 

\subsection{Overview}

Our goal is to improve the interpretability of DNNs without losing efficiency. By designing proper learning objective, we expect to learn the hidden features with two properties---discriminability and interpretability. Discriminability defines the ability that the features can distinguish different inputs and predict corresponding outputs. Interpretability measures the extent that human users can understand and manipulate the learned features. These two properties are often contradictory in DNNs. According to the fundamental bias-variance tradeoff, a complex DNN can be highly competitive in prediction performance but its hidden features are often too abstract to be understandable for humans. 
On the other hand, a simple DNN can lead to more interpretable features, but it may degrade the performance. 
In order to break this dilemma, we introduce extra semantic information to guide the learning process. 
In this paper, we will concentrate on the video captioning task~\cite{venugopalan2014translating}, although the similar ideas can be generalized to other scenarios. 

Specifically, the video captioning task aims to automatically describe video content with a complete and natural sentence. Recent works have demonstrated that video captioning can benefit from the discovery of multiple semantics, including obejcts, actions, relationships and so on. Liu \etal.~\cite{liu2017hierarchical,liu2016benchmarking} proposed one original method for joint human action modeling and grouping, which can provide comprehensive information for video caption modeling and explicitly benefit understanding what happens in the given video. As a video is more than a set of static images, in which there are not only the static objects but also the temporal relationships and actions, video analysis often requires more complex network architectures. For example, some works have shown the effectiveness of DNNs on video analysis~\cite{ballas2015delving,yu2015video} when stacking a hierarchical RNN on top of some CNN layers. Such a complex network makes it more challenging to learn interpretable hidden features and hinders the interaction between the models and human users. To address this issue, we propose a novel technique to improve the interpretability of the learned features by leveraging the latent topics extracted from video descriptions.

The overall framework is shown in Fig.~\ref{fig:framework}, which consists of an attentive encoder-decoder network for video caption generation and an interpretive loss to guide the learning towards semantically meaningful features.

Formally, in the training set, each video $\mathbf{x}$ has $n$ sample frames along with a set of $N_d$ descriptions $\mathbf{Y} = \{\mathbf{y}^1,\mathbf{y}^2,...,\mathbf{y}^{N_d}\}$. For each $\mathbf{y} \in \mathbf{Y}$, let $(\mathbf{x},\mathbf{y})$ denote a training video-description pair, where $\mathbf{y} = \{y_1,y_2,...,y_{N_s}\}$ is a description with $N_s$ words. 
We first transform the input $\mathbf{x}$ into a set of $D_v$-dimensional hidden features $V = \{\mathbf{v}_1, ..., \mathbf{v}_n\}$ by using an encoder network. Then, the hidden features are decoded to generate the description $\mathbf{y}$. We define the task-specific loss as the negative log-likelihood of the correct description
\begin{equation}
L_T(\mathbf{x}, \mathbf{y}) = -\log p(\mathbf{y}|\mathbf{x}).
\end{equation}
We parse the text descriptions $\mathbf{Y}$ to get a semantically meaningful representation (\ie, a topic representation in this paper), which is denoted as $\mathbf{s}$. Then, we introduce an interpretive loss $L_I(V, \mathbf{s})$ to measure the compliance of the learned features with respect to the semantic representation $\mathbf{s}$. Putting together, we define the overall objective function as 
\begin{equation}
L(\mathbf{x},\mathbf{y},\mathbf{s}) = -\log p(\mathbf{y}|\mathbf{x}) + \lambda L_I(V, \mathbf{s}).
\end{equation}
The tradeoff between these two contradictory losses is captured by a balancing weight $\lambda$. An interpretation system with high-quality can be realized based on an appropriate $\lambda$, which can be obtained using the validation set.  

After training (See the experimental section for details), we use the prediction difference maximization algorithm to interpret the learned features of each neuron by a topic. Below, we elaborate each part. 

\subsection{Attentive Encoder-Decoder Framework}
We adopt an attentive encoder-decoder framework similar to~\cite{yao2015describing} for video captioning. The attention mechanism is used to let the decoder selectively focus on only a small subset of frames at a time.

A key difference from previous works~\cite{venugopalan2014translating,yao2015describing,pan2015jointly} which use CNN features as video representations is that we stack a bi-directional LSTM model~\cite{schuster1997bidirectional} on top of a CNN model to characterize the video temporal variation in both input directions. Such an encoder network makes the vector representation $\mathbf{v}_i$ of the \textit{i}-th frame capture temporal information, and thus the interpretive loss (defined later in Eq.~\ref{eq:interpretive}) lets the internal neurons learn to detect  latent topics in the video. So the learned features are more likely to be both discriminative and interpretable.

To generate the description sentences, we use an LSTM model as the decoder. At each time step, the input to the LSTM decoder can be represented by $[\mathbf{y}_{t-1},\phi_t(V)]$, where $\mathbf{y}_{t-1}$ is the previous word and $\phi_t(V)$ is the dynamic weighted sum of temporal feature vectors
\begin{equation}
\phi_t(V) = \displaystyle\sum_{i=1}^{n} \alpha_i^t\mathbf{v}_i.
\end{equation}
The attention weight $\alpha_i^t$ reflects the importance of the \textit{i}-th temporal features at time step \textit{t}~\cite{yao2015describing}, which is defined as % which can be obtained by
\begin{equation}
\alpha_i^t = \frac{\exp(\mathbf{w}_a\tanh(\mathbf{U}_a\mathbf{h}_{t-1} + \mathbf{T}_a\mathbf{v}_i + \mathbf{b}_a))}{\sum_{j=1}^{n} \exp(\mathbf{w}_a\tanh(\mathbf{U}_a\mathbf{h}_{t-1} + \mathbf{T}_a\mathbf{v}_j + \mathbf{b}_a))},
\end{equation}
where $\mathbf{w}_a$, $\mathbf{U}_a$, $\mathbf{T}_a$ and $\mathbf{b}_a$ are the parameters that are jointly estimated with the other parameters. We adopt the same strategy as \cite{xu2015show} to initialize the memory state and hidden state as
\begin{equation}
\left[\!
\begin{array}{c}
\mathbf{c}_0 \\
\mathbf{h}_0
\end{array}
\!\right] = 
\left[\!
\begin{array}{c}
f_{init,c} \\
f_{init,h}
\end{array}
\!\right]
(\frac{1}{n}\displaystyle\sum_{i=1}^{n} \mathbf{v}_i),
\end{equation}
where $f_{init,c}$ and $f_{init,h}$ are both multilayer perceptions, which can  also be jointly estimated.%\junz{how to estimate the parameters of the MLP?}

At each time step, we use the LSTM hidden state $\mathbf{h}_t$ to predict the following word, and define a probability distribution over the set of possible words by using a softmax layer
\begin{equation}
\mathbf{p}_t = \mathrm{softmax}(\mathbf{W}_p[\mathbf{h}_t,\phi_t(V),\mathbf{y}_{t-1}] + \mathbf{b}_p).
\end{equation}
Therefore, we can predict the next word based on such probability distribution until the end sign is emitted. The log-likelihood of the sentence is therefore the sum of the log-likelihood over the words
\begin{equation}
\log p(\mathbf{y}|\mathbf{x}) = \displaystyle\sum_{t=1}^{N_s} \log p(\mathbf{y}_t|\mathbf{y}_{<t},\mathbf{x};\theta),
\end{equation}
where $\theta$ are the parameters of the attentive encoder-decoder model. %\junz{which model?}.

\subsection{Interpretive Loss}
The above architecture for video captioning incorporates both CNN and RNN to encode the spatial and temporal information. The complex architecture makes internal neurons learn more abstract features than a single CNN or RNN, and these features are typically hard to interpret by human users. 
To improve the interpretability, we introduce an interpretive loss, which makes the neurons learn to detect semantic attributes in the text descriptions. For humans, it is natural and easy to understand a concept in text descriptions. 

In our method, instead of using the raw description data which can be very sparse and high-dimensional vectors (\eg, in bag-of-words or tf-idf format), we adopt a topic model to learn a semantic representation. As proven in previous work~\cite{fei2005bayesian,cao2007spatially}, topic models can extract semantically meaningful concepts (or themes) that are useful for visual analysis tasks. Furthermore, compared to the raw text descriptions, the representations by topic models can better capture the global statistics in a corpus as well as synonymy and polysemy~\cite{blei2003latent}. 
Here, we adopt the most popular topic model, \ie, Latent Dirichlet Allocation (LDA)~\cite{blei2003latent}, which has been applied to image/video/text analysis tasks~\cite{fei2005bayesian,cao2007spatially,boyd2007topic,zhu2012medlda}.
Specifically, LDA is a hierarchical Bayesian model, in which each document is represented as a finite mixture over topics and each topic is characterized by a distribution over words.
In our case, we concatenate all of the single descriptions in $\mathbf{Y}$ together to form a ``document''.
Here, we adopt WarpLDA~\cite{chen2015warplda} to efficiently estimate the parameters for an $N_t$ topics LDA model, and set $N_t$ to $100$ in experiments. The top words from the learned topics are illustrated in Table.~\ref{tab:LDA}. We can see that  each topic has a good correspondence to a meaningful semantic attribute. 
\begin{table}
\begin{center}
\begin{tabular}{|c|c|}
\hline
\textbf{people} & people, group, men, line, crowd\\
\hline
\textbf{woman} & woman, lady, women, female, blond\\
\hline
\textbf{man} & man, guy, unique, kind, bare\\
\hline
\textbf{dance} & dancing, dance, stage, danced, dances\\
\hline
\textbf{walk} & walking, walks, race, turtle, walk\\
\hline
\textbf{eat} & eating, food, eats, eat, ate \\
\hline
\textbf{play} & playing, plays, play, played, instrument\\
\hline
\textbf{field} & grass, field, yard, run, garden\\
\hline
\textbf{dog} & dog, tail, barking, wagging, small\\
\hline
\textbf{cat} & cat, licking, cats, paws, paw\\
\hline
\end{tabular}
\end{center}
\vspace{-1ex}
\caption{Sampled latent topics with their high-probability words. We have named the topics according to these words.}
\label{tab:LDA}\vspace{-2ex}
\end{table}

After training, we parse each description document to get the latent topic representation, from which the words are generated. We encode the topic representation for a video as a binary vector $\mathbf{s} = [t_1, t_2, ...,t_{N_t}] \in \{0, 1\}^{N_t}$, whose \textit{i}-th element $t_i$ is set to $1$ when \textit{i}-th topic occurs in the descriptions, and $0$ otherwise.
This vector can be obtained by running the Gibbs sampler in WarpLDA. We use a binary vector here rather than a real-valued vector denoting the average probability of each topic, because it provides an easy way to interpret the learned features of each neuron by a topic, when applying the prediction difference maximization algorithm described in Section.~\ref{sec:PDM}. 

Given the topic representations, we define the interpretive loss as
\begin{equation}
L_I(V, \mathbf{s}) = \|f(\frac{1}{n} \displaystyle\sum_{i=1}^{n} \mathbf{v}_i) - \mathbf{s}\|_2^2,
\label{eq:interpretive}
\end{equation}
where $f : D_v \to N_t$ is an arbitrary function mapping video features to topics. 
This formulation can be individually viewed as a multi-label classification task, where we predict topics given a set of video features. The choice of the function $f$ is also a tradeoff between interpretability and task performance. A complex function with a large number of parameters will increase the interaction among different neurons, leading to hard-to-interpret features again. On the contrary, a too simple function will limit the discriminability of the learned features. For example, an identity mapping will turn the hidden features into a replica of topics, which may degrade the performance for caption generation. Here, we adopt a two-layer perception as $f$. To avoid overfitting, we use ``mean pooling'' features over all frames as input. We use $l_2$-norm to define interpretive loss because it's simple and effective to build a correspondence between neurons and topics and it performs well in practice. We will see in Section.~\ref{visual} how the interpretive loss help to learn interpretable features.

\subsection{Prediction Difference Maximization}
\label{sec:PDM}
To analyze the correspondence between neurons and topics and semantically interpret the learned features, we propose a prediction difference maximization algorithm, which is similar with a concurrent and independent work~\cite{zintgraf2017visualizing}, to represent the learned features of each neuron by a topic. 
This method is different from the activation maximization methods~\cite{erhan2009visualizing,nguyen2016multifaceted}, where they aim to find the input patterns (\eg, image patches) that maximally activate a given neuron.
The reason why we do not use activation maximization is that some neurons represent temporal actions (\eg, playing, eating), which cannot be represented by static image patches.

Specifically, in a video with topic representation $\mathbf{s}$, for each topic \textit{i} that $t_i = 1$ and $i \in [1,...,N_t]$, we expect to find a neuron $j_i^{\ast}$ that
\begin{equation}
j_i^{\ast} = \argmax_j([f(\mathbf{v})]_i - [f(\mathbf{v}_{\backslash j})]_i),
\label{equ:PDM}
\end{equation}
where $\mathbf{v} = \frac{1}{n} \sum_{i=1}^{n} \mathbf{v}_i$ are the average video features, $\mathbf{v}_{\backslash j}$ denotes the set of all input features except that the $j$-th neuron is deactivated (set to zero) and $[f(\cdot)]_i$ is the \textit{i}-th element of the prediction $f(\cdot)$. 

The purpose of Eq.~\ref{equ:PDM} is to find a neuron which contributes most to predicting a topic occurred in the video. We can consider that the identified neuron $j_i^{\ast}$ ``prefers'' topic $i$, which can then represent the learned features of $j_i^{\ast}$. After we go through all the videos in the training set, we can find one or more neurons associated with each topic. 
Note that previous work has shown that a neuron may respond to different facets~\cite{nguyen2016multifaceted}, which is also true in our case, that is, a neuron may prefer different topics. Here, we only choose one for simplicity %, with which the neuron has higher activation, 
to represent the learned features of the given neuron. 

\section{Experimental Results}

\subsection{Experimental Settings}
\textbf{Dataset:} We use the YouTubeClips~\cite{chen2011collecting} dataset, which is well suited for training and evaluating an automatic video captioning model. The dataset has $1,970$ YouTube clips with around $40$ English descriptions per video. The video clips are open-domain, containing a wide range of daily subjects like sports, animals, actions, scenarios, etc. Following~\cite{venugopalan2014translating}, we use $1,200$ video clips for training, $100$ video clips for validation and $670$ video clips for testing.

\textbf{Training:} In the attentive encoder-decoder framework, we select $n = 28$ equally sampled frames in each video and feed each frame into GoogLeNet~\cite{szegedy2015going} to extract a $1024$ dimensional frame-wise representation from the $pool5/7\times7\_s1$ layer. The parameters of GoogLeNet are fixed during training.

The overall objective function for caption generation is
\begin{equation*}
L = \frac{1}{N} \displaystyle\sum_{k=1}^{N} \Big(\lambda \|f(\mathbf{v}^k) - \mathbf{s}^k\|_2^2 - \displaystyle\sum_{t=1}^{N_s^k} \log p(\mathbf{y}_t^k|\mathbf{y}_{<t}^k,\mathbf{x}^k)\Big),
\end{equation*}
where there are $N$ training video-description pairs $(\mathbf{x}^k, \mathbf{y}^k)$. $\mathbf{v}^k = \frac{1}{n} \sum_{i=1}^{n} \mathbf{v}_i^k$ are the average video features and $\mathbf{s}^k$ is the topic representation for video $\mathbf{x}^k$. 

We use Adadelta~\cite{zeiler2012adadelta} to jointly estimate the model parameters for bi-directional LSTM of the encoder, attentive LSTM of the decoder and two-layer perception of $f$. After training, we apply the prediction difference maximization algorithm to interpret the learned features of each neuron.

\textbf{Baseline:} In the experiment, we call our model LSTM-I which jointly models the interpretability of the learned features and video captioning. The hyperparameter $\lambda$ is set to 0.1 by maximizing the performance on the validation set. To compare the results, we also test a baseline model named LSTM-B without interpretive loss---it is only optimized with respect to the sum of negative log-likelihood over the words. These two models have the same encoder-decoder architecture. 

\subsection{Feature Visualization}
\label{visual}
\begin{figure}[t]
\begin{center}
  \includegraphics[width=1.0\linewidth]{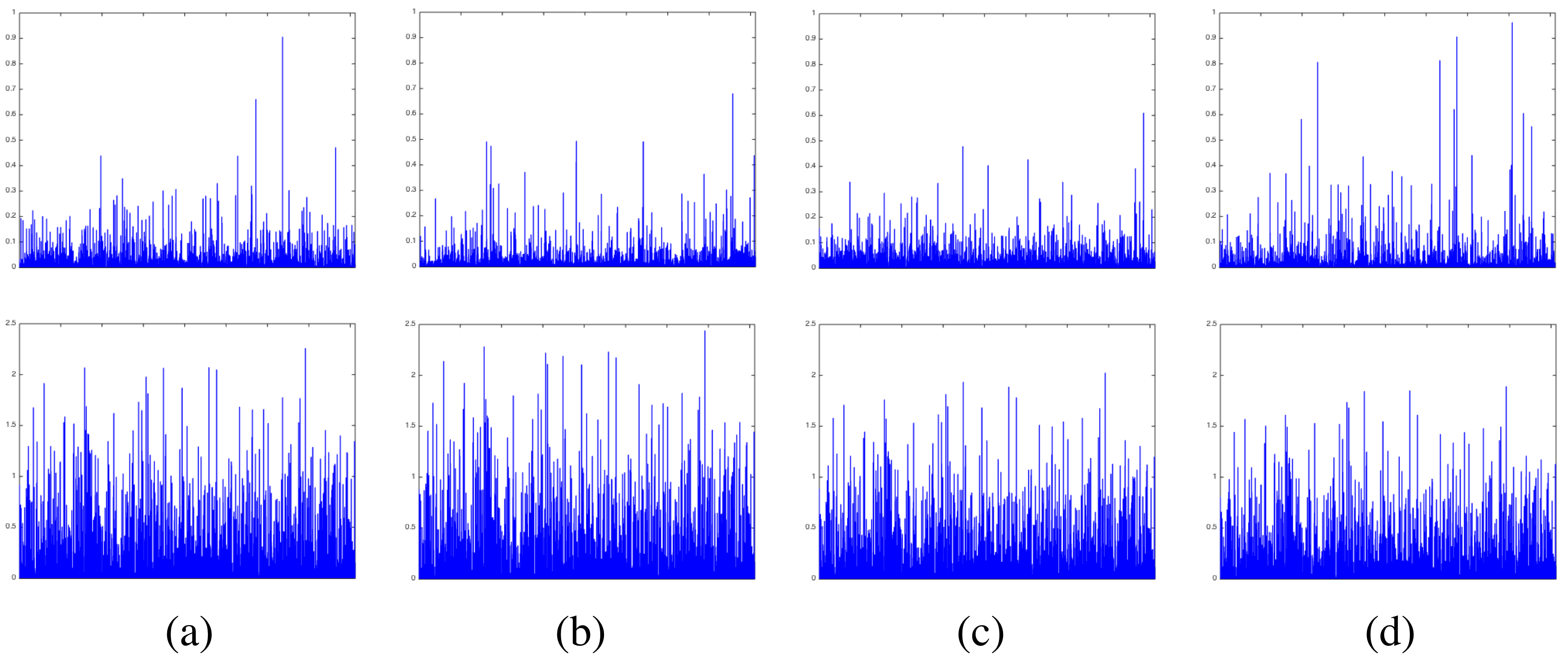}
\end{center}\vspace{-2ex}
   \caption{Examples of learned video representations using LSTM-I model (top) and LSTM-B model (bottom). Each histogram indicates an average of activations of a subset of videos, which have the same topic. (a) representations for topic ``dog''; (b) representations for topic ``girl''; (c) representations for topic ``walk''; (d) representations for topic ``dance''. The top words for topics are shown in Table.~\ref{tab:LDA}.}
\label{fig:neuron}\vspace{-2ex}
\end{figure}

\begin{figure*}
\begin{center}
\includegraphics[width=0.95\linewidth]{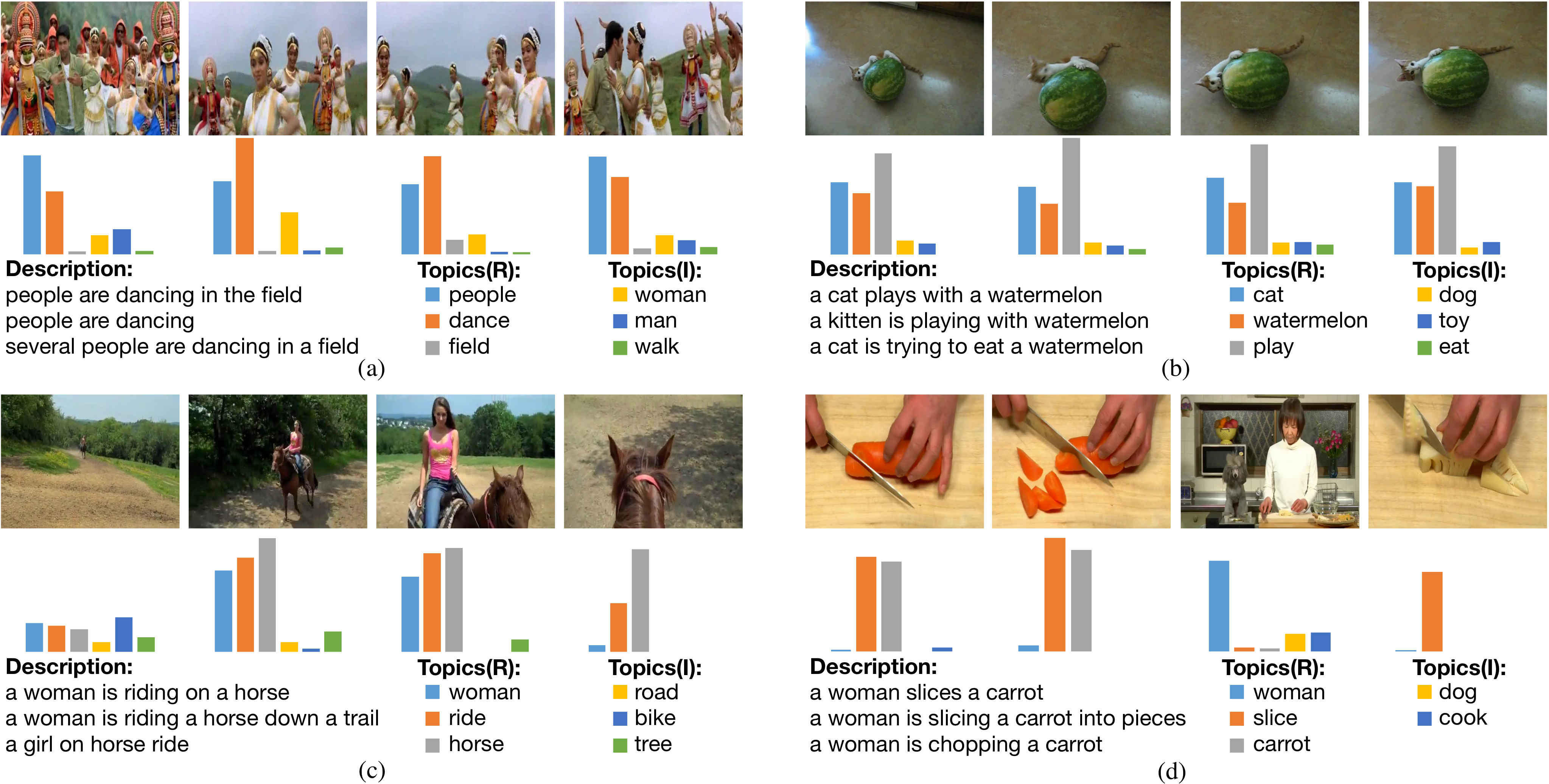}
\end{center}\vspace{-2ex}
   \caption{Neuron activations with respect to relevant and irrelevant topics in sampled videos. Topics(R) are relevant topics extracted from video descriptions. Topics(I) are irrelevant topics which are unimportant or easily confusing topics. We plot the activations of one neuron related to each topic through time. 
   %\junz{this is not a typical way to represent topics; normally a topic is represented by a vector of high-probability words.}
   }
\label{fig:activation}
\vspace{-2ex}
\end{figure*}

We visualize the learned representations of test videos in Fig.~\ref{fig:neuron}. The top and bottom rows show the results of LSTM-I and LSTM-B, respectively. We randomly choose some topics, and for each topic, we find a subset of videos containing this topic and plot the neuron activations by averaging the features from these videos. It can be seen that the entries of LSTM-I representations are very peaky at some specific neurons, indicating a strong correspondence between topics and neurons. Therefore we can rely on the prediction difference maximization algorithm to represent the learned features of neurons by the corresponding semantic topics. The interpretability of the learned features in LSTM-I is better than that in LSTM-B because the correspondence between neurons and topics decouples the interaction of different neurons, which makes human users easily understand and manipulate video features (See Section.~\ref{sec:human-in-the-loop}).

To further visualize the variation of the learned features for a video through time, we randomly choose some videos for presentation as shown in Fig.~\ref{fig:activation}. For each video, we select some relevant topics existing in the descriptions and some irrelevant topics which are unimportant or easily confusing\footnote{A video demo is available at \url{http://ml.cs.tsinghua.edu.cn/~yinpeng/papers/demo-cvpr17.mp4}}. We have named the topics according to their high-probability words (See Table.~\ref{tab:LDA} for the top words with respect to topics). We plot the activations of one neuron associated with each topic. We can see that the neurons associated with the salient topics in videos have high activations through time, which suggests the strong compliance between neuron activations and video contents. 

It should be noted that the relevant topics are mapped from average video features in Eq.~\ref{eq:interpretive}, so the associated neurons may not be activated in every frame. For example, in Fig.~\ref{fig:activation} (c), the neuron with respect to topic ``horse'' is not activated in the first frame, but the average video features can be mapped to predict topic ``horse'' correctly. The fact is also true in Fig.~\ref{fig:activation} (d), where the neuron with respect to topic ``woman'' is only activated in the third frame. It proves that the neurons are able to detect corresponding topics when they appear without severe overfitting.

\subsection{Performance Comparison}

\begin{table}
\begin{center}
\begin{tabular}{|l|c|c|}
\hline
 Model & BLEU & METEOR \\
\hline\hline
LSTM-B (GoogLeNet) & 0.416 & 0.295\\
LSTM-I (GoogLeNet) & \textbf{0.446} & \textbf{0.297}\\ 
\hline
LSTM-YT (AlexNet)~\cite{venugopalan2014translating} & 0.333 & 0.291\\
S2VT (RGB + Optical Flow)~\cite{venugopalan2015sequence} & - & 0.298\\
SA (GoogLeNet)~\cite{yao2015describing} & 0.403 & 0.290\\
LSTM-E (VGG)~\cite{pan2015jointly} & 0.402 & 0.295\\
h-RNN (VGG)~\cite{yu2015video} & 0.443 & \textbf{0.311}\\
\hline
SA (GoogLeNet + C3D)~\cite{yao2015describing} & 0.419 & 0.296\\
LSTM-E (VGG + C3D)~\cite{pan2015jointly} & 0.453 & 0.310\\
h-RNN (VGG + C3D)~\cite{yu2015video} & 0.499 & 0.326\\
\hline
\end{tabular}
\end{center}
\vspace{-1ex}
\caption{BLEU and METEOR scores comparing with the state-of-the-art results of description generation on YouTubeClips dataset.}
\label{tab:performance}\vspace{-2ex}
\end{table}

To validate whether the interpretability of the learned features will affect the task performance, we test the quality of the generated sentences measured by BLEU~\cite{papineni2002bleu} and METEOR~\cite{denkowski2014meteor} scores, which compute the similarity between a hypothesis and a set of references. In the first block of Table.~\ref{tab:performance}, we compare the performance with the baseline model. We can see that LSTM-I significantly outperforms the baseline LSTM-B in BLEU score and achieves better performance in METEOR score. 
%\junz{where are the results? how much improvement?}. 
These results suggest that our model benefits from the proper way of incorporating external semantic information into the training process, which makes the features capture more useful temporal information (\eg, actions, relationships) and thus generates more accurate descriptions. So the proposed LSTM-I with better interpretability also helps to improve the performance for captioning.

To fully evaluate the performance on the video captioning task, we compare our approach with five state-of-the-art methods, namely, LSTM-YT~\cite{venugopalan2014translating}, S2VT~\cite{venugopalan2015sequence}, SA~\cite{yao2015describing}, LSTM-E~\cite{pan2015jointly}, and h-RNN~\cite{yu2015video}. %\junz{add some brief description of the key features for each baseline?}
LSTM-YT translates videos to descriptions with a single network using mean pooling of AlexNet~\cite{krizhevsky2012imagenet} features over frames; S2VT directly maps a sequence of frames to a sequence of words; SA incorporates a soft attention mechanism into the encoder-decoder framework; LSTM-E considers the relationship between the semantics of the entire sentence and video content by embedding visual-semantic; and h-RNN exploits the temporal dependency among sentences in a paragraph by a hierarchical-RNN framework.

We only use 2-D CNN features for simplicity in this work. For fair comparison, we also show the extensive results of SA, LSTM-E and h-RNN which only incorporate 2-D CNN features. By comparing our baseline method LSTM-B to SA, which only uses a single CNN model as encoder and a similar decoder architecture, we can see that our baseline model achieves higher BLEU and METEOR scores, suggesting that the bi-directional LSTM can help us capture temporal variation and lead to better video representations. On the other hand, our LSTM-I achieves state-of-the-art performance and gets higher BLEU score than other methods. These results further demonstrate the effectiveness of our method, and we can conclude that integrating the interpretability of latent features by leveraging semantic information during training is a feasible way to simultaneously improve interpretability and achieve better performance.

\subsection{Video Action Recognition}

We also demonstrate the generalization ability of the learned interpretable features. Specifically, 
interpretable features usually contain more general information than the features learned by optimizing a task-specific objective, because task-specific features may overfit the particular dataset, while interpretable features reach a good balance between task-specific fitting and generalization.
We test this hypothesis by examining a transfer learning task---we evaluate the performance on video action recognition by transferring the learned features from video captioning. 

We use the UCF11 dataset \cite{Liu2009recognizing}, a YouTube action dataset consisting of $1600$ videos and $11$ actions, including basketball shooting, biking, diving, golf swinging, horse riding, soccer juggling, swinging, tennis swinging, trampoline jumping, volleyball spiking, and walking. Each video has only one action associated with it. We randomly choose $800$ videos for training and $800$ videos for testing.
We use the same encoder architecture as in the video captioning task. The decoder is a two-layer perception and we minimize the cross-entropy loss. 

\begin{table}
\begin{center}
\begin{tabular}{|l|c|c|c|}
\hline
 Model & LSTM-B & LSTM-I & LSTM-R \\
\hline
Accuracy(\%) & 88.50 & \textbf{91.13} & 90.13 \\
\hline
\end{tabular}
\end{center}
\vspace{-1ex}
\caption{Video action recognition performance of different models.}\vspace{-2ex}
\label{tab:action}
\end{table}

Table.~\ref{tab:action} presents the results, where we adopt three model variants. We do not compare with other state-of-the-art action recognition models due to the lack of standard train-test splits. In LSTM-B and LSTM-I, the parameters of the encoder are fixed, which are initialized with the trained captioning model. We can consider that the features for action recognition are extracted from the trained captioning encoder. We only optimize the parameters of the two-layer decoder. LSTM-R has the same architecture but all the parameters are initialized randomly and then optimized. 

We can see that LSTM-I achieves higher accuracy than LSTM-B, which verifies our hypothesis that the interpretable features contain more general information than the task-specific features and lead to higher performance in other tasks. Another fact is that LSTM-I outperforms LSTM-R, which indicates that the interpretable features learned by captioning task are more effective than the features learned by action recognition. This is because that the interpretive loss makes the neurons learn to detect semantic attributes in LSTM-I model and these features are general and transferable for other tasks.

\section{Human-in-the-loop Learning}
\label{sec:human-in-the-loop}

An important advantage of the interpretable features is that they provide a natural interface to get human users involved in the learning process, and make them understand how the system works, what is going wrong and how to correct errors (if any). 
Although previous works~\cite{zeiler2014visualizing,liu2016towards} have provided some applications on human interaction and architecture selection, they still need expert-level users to join the procedure because non-expert users get little insight about potential problems. They also lack a human-in-the-loop learning procedure, which helps models integrate human knowledge into the training process to refine their shortcomings.
Here, we present an easy way to allow a human-in-the-loop learning procedure by clearly understanding the learned features without requiring expert-level experience of human users. In our case, when the model outputs an inaccurate description, human users only need to provide the missing topics. The human-in-the-loop learning procedure can diagnose potential problems in the model and refine the architecture, so the similar errors will never occur in future unseen data.

Specifically, the human-in-the-loop learning procedure can be divided into two steps---activation enhancement and correction propagation. First, when it receives a topic $t$ from human users for an inaccurate output, it retrieves a set of neurons associated with $t$, which are already found by the prediction difference maximization algorithm. For these neurons, it adds the average activations of them in a subset of training videos containing topic $t$ and turns the original features $\mathbf{v}$ to $\mathbf{v}^{\ast}$.  The purpose of activation enhancement is to let the neurons associated with the missing topic have higher activations, so the decoder propobably maps the new features to more accurate descriptions. Second, to generalize the specific error to future unseen data, we use the correction propagation to fine-tune the parameters of the encoder. We expect to let the encoder learn to generate $\mathbf{v}^{\ast}$ instead of $\mathbf{v}$, so we minimize
\begin{equation}
\label{eq:human}
L_{human} = \|\mathbf{v}^{\prime} - \mathbf{v}^{\ast}\|_2^2 + \mu \|\theta^{\prime} - \theta\|_2^2,
\end{equation}
where $\mathbf{v}^{\prime}$ are the outputs of the refined encoder, $\theta$ and $\theta^{\prime}$ are the parameters of the original and the refined encoders, respectively. The first term aims to let features $\mathbf{v}^{\prime}$ approximate the optimal features $\mathbf{v}^{\ast}$, and the second term forces the model to have little variations. The balancing weight $\mu$ makes the refined model not overfit to this sample.

\begin{figure}[t]
\begin{center}
\includegraphics[width=1.0\linewidth]{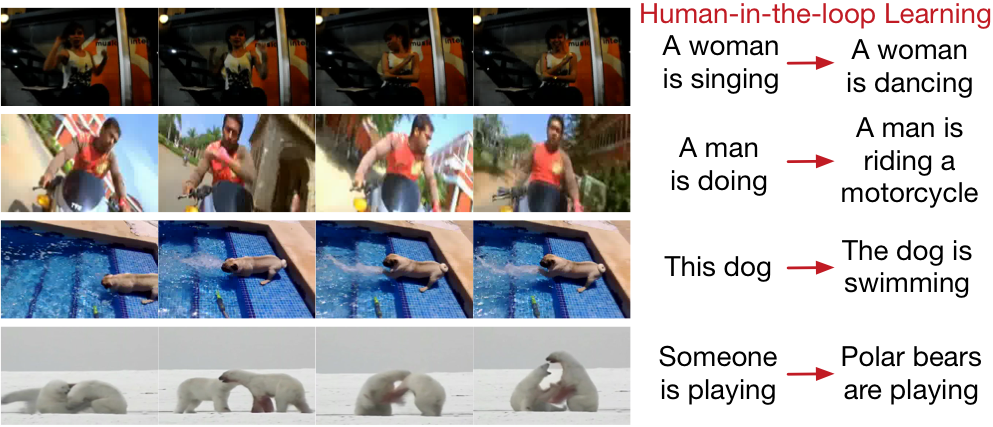}
\end{center}\vspace{-2ex}
   \caption{We show the second half (unseen part) of four videos and the predicted captions before and after refining the model. By providing the missing topics (``dance'', ``motorcycle'', ``swim'' and ``polar bear'') for the first half of these four videos and refining the model, the predictions for the second half are more accurate.}
\label{fig:interaction}
\vspace{-2ex}
\end{figure}

In our experiments, it's hard to find a similar error occurred in two videos in the test set due to its small size and rich diversity. So we find $20$ videos with inaccurate predictions in the test set and split each of them into two parts. We optimize Eq.~\ref{eq:human} using the first half of each video and use the second half as future unseen data to test. We get more accurate captions for 17 videos (second half), which can capture the missing topics. Fig.~\ref{fig:interaction} shows four cases. Taking the polar bears video for example, the model doesn't capture the salient object ``polar bear'' for every parts of the video. By providing the missing topic and refining the model using the first half, the model can accurately capture ``polar bear'' and produce more accurate captions for the second half. It proves that the model learns to solve its potential problems with the aid of human users.

We also examine whether this procedure could affect the overall performance. We test the performance of the new model after refining for these $20$ videos in turn. We get BLEU score $0.449$ and METEOR score $0.298$, which are sightly better than those in Table.~\ref{tab:performance} to prove that refined model does not affect the captioning ability while it makes more accurate predictions for the selected $20$ videos.

\section{Conclusions}

In this work, we propose a novel technique to improve the interpretability of deep neural networks by leveraging human descriptions. We base our technique on the challenging video captioning task. In order to simultaneously improve the interpretability of the learned features and achieve high performance, we extract semantically meaningful topics from the corpus and introduce an interpretive loss during the training process. To interpret the learned features in DNNs, we propose a prediction difference maximization algorithm to represent the learned features of each neuron by a topic. We also demonstrate a human-in-the-loop training procedure which allows humans to revise false predictions and help to refine the network. 

Experimental results show that our method achieves better performance than various competitors in video captioning. The learned features in our method are more interpretable than opaque models, which can be transferred to video action recognition. Several examples prove the effectiveness of our human-in-the-loop learning procedure.

{\small
\bibliographystyle{ieee}
\bibliography{egbib}
}

\end{document}